\documentclass[runningheads]{llncs}
\usepackage[T1]{fontenc}
\usepackage{graphicx}
\usepackage{amsmath}
\usepackage{amssymb}
\usepackage{booktabs,multirow}
\usepackage{xcolor}
\usepackage{tcolorbox} 
\usepackage{wasysym}
\usepackage{tikzsymbols}
\usepackage{fontawesome5}
\usepackage{graphicx,verbatim}
\usepackage{orcidlink}
\usepackage{hyperref}
\begin{document}
\title{Parameter-Efficient Self-Supervised Adaptation for EEG-FM under Fixed Computational Budgets}
\titlerunning{Parameter-Efficient SSL Adaptation for EEG-FM}
%
\author{Meghal Dani\inst{1,3}\orcidID{0009-0007-1249-3630} \and
Stefanie Liebe\inst{2,3,4}\orcidID{0000-0003-2873-2943} }
%
\authorrunning{M. Dani et al.}
%
\institute{University of Tübingen, Germany \and
Dept. of Neurology and Epileptology, University Clinic Tübingen, Germany \\ \and 
Hertie Institute for AI in Brain Health (Hertie AI), Tübingen, Germany\\ \and
Hertie Institute for Clinical Brain Research, Tübingen, Germany \\ 
\email{meghal.dani@uni-tuebingen.de}}
\maketitle              
\begingroup
\renewcommand\thefootnote{}\footnotetext{Accepted at MICCAI 2026 (Strasbourg, France). Preprint of the submitted version.}%
\addtocounter{footnote}{-1}\endgroup
\begin{abstract}
EEG foundation models pretrained via self-supervised learning promise transferable representations, but their generalization remains limited, especially across diverse clinical datasets. Full fine-tuning is impractical for resource-constrained clinical settings due to high computational requirements. In this work, we investigate whether parameter-efficient self-supervised adaptation, updating only 9\% of parameters suffices to align representations to target tasks. We evaluate our method on two state-of-the-art models with different pretraining objectives: BIOT (contrastive) and CBraMod (masked reconstruction), and evaluate on three clinical EEG datasets for abnormality detection (TUAB), event classification (TUEV), and seizure detection (CHB-MIT) under both in-distribution and out-of-distribution conditions. SSL adaptation yields consistent gains over linear probing, up to $20\times$ AUCPR. Under a fixed compute budget, peak performance requires only 20--50\% of available unlabeled data. Critically, when total window count is fixed, performance remains invariant to patient count, suggesting that performance is dependent on overall temporal window diversity only. Our findings demonstrate that parameter-efficient adaptation enables effective deployment of EEG Foundation models (EEG-FM) with minimal computational overhead and data collection burden. \\ Code available at: \href{https://github.com/c3n-group/efficient-eeg-adapt}{https://github.com/c3n-group/efficient-eeg-adapt} 

\keywords{EEG  \and Self Supervised Learning \and Foundation Models.}
\end{abstract}
\section{Introduction}
Electroencephalography (EEG) is the gold standard for non-invasive neurological monitoring and still heavily relies on labor-intensive manual expert inspection~\cite{mushtaq2024one,bitar2024utility}. Deep learning algorithms using supervised training have addressed automating this process~\cite{van2019detecting,amin2019cognitive,alhussein2019eeg,gemein2020machine}. Although unlabeled EEG recordings are abundant at clinical sites, annotated data sets are rare, limiting these approaches. 

In contrast, Foundation models (FMs), neural networks pretrained on massive unlabeled corpora via self-supervised learning (SSL), hold promise for not requiring any expert annotations. A first generation of EEG foundation models (EEG-FMs) has emerged~\cite{kuruppu2026eeg,yang2023biot,jiang2024labram,jiang2025neurolm,wang2025cbramod,wang2024eegpt}, demonstrating that broad pretraining on unlabeled EEG can yield transferable representations. Most models segment raw multichannel EEG into fixed-length time-series patches per channel, tokenized into embeddings that preserve both spatial and temporal structure, and processed by transformers. Their SSL objectives fall into two families: contrastive learning (e.g. BIOT~\cite{yang2023biot}) and masked reconstruction (e.g. CBraMod~\cite{wang2025cbramod}).

Despite this progress, deploying a pretrained EEG-FM to a clinical site requires bridging distribution shifts caused by differences in recording hardware, channel montages, and sampling rates. Retraining from scratch is impractical given the annotation costs, motivating parameter-efficient adaptation strategies. REMEDIS~\cite{azizi2023robust} established that SSL-based adaptation of pretrained models outperforms both direct transfer and full fine-tuning in medical imaging, particularly under limited labeled data. For EEG, concurrent work has been explored using graph-based adapters~\cite{suzumura2024graph}, while recent EEG-FMs such as LaBraM~\cite{jiang2024labram} and EEGPT~\cite{wang2024eegpt} have scaled pretraining data, though with diminishing returns. Specifically, NeuroLM~\cite{jiang2025neurolm}, trained on roughly ten times more data than LaBraM, achieves comparable downstream performance. This raises a fundamental question: \emph{how much unlabeled target-domain data is actually needed for reliable EEG-FM adaptation?}

We address this by studying parameter-efficient SSL adaptation in a setting that mirrors clinical deployment: adapt a pretrained EEG-FM using only unlabeled target-domain EEG, then evaluate representation quality via frozen-encoder linear probing. We probe two EEG-FMs with different pretraining objectives, BIOT (contrastive) and CBraMod (masked reconstruction), updating only the final encoder layer while preserving each model's original SSL objective. To make data-efficiency conclusions actionable, we evaluate our adaptation under a normalized fixed-compute protocol~\cite{al2024pretrainingtorr_ssl} that holds total computation constant. Across three clinical EEG datasets covering both in-distribution (ID) and out-of-distribution (OOD) conditions, we find that: (1) lightweight SSL adaptation substantially improves linear probe representations with gains up to $20{\times}$ in AUCPR on rare-event tasks, (2) near-peak performance is achieved using only a fraction ($20-50$\%) of available unlabeled data, (3) the number of total temporal windows seen, rather than the number of unique patients, impacts adaptation performance under a fixed compute budget.

\section{Method}
\subsection{Parameter Efficient SSL Adaptation}
Our pipeline consists of two stages: (i) SSL-based representation learning for parameter-efficient adaptation, and (ii) linear probing for downstream evaluation.

\textbf{Self Supervised Learning Pretraining}: 
Let $f_{\theta}$ denote an EEG foundation model encoder that maps an input EEG segment $\mathbf{x} \in \mathbb{R}^{C \times T}$ (with $C$ channels and $T$ time steps) to a representation $\mathbf{h}=f_{\theta}(\mathbf{x}) \in \mathbb{R}^d$. To effectively adapt the foundation model under resource-constrained settings, we adopt a parameter-efficient fine-tuning strategy: we partition the full parameter set into frozen lower layers $\theta_{\mathrm{frozen}}$ and a trainable subset $\theta_{\mathrm{adapt}}$ corresponding to the final encoder layer, such that $\theta = \{\theta_{\mathrm{frozen}},\, \theta_{\mathrm{adapt}}\}$. We preserve the original SSL objective of each model, i.e., the masked reconstruction for CBraMod (with masking ratio 0.5~\cite{wang2025cbramod}), contrastive learning for BIOT and optimize the loss with respect to $\theta_{\mathrm{adapt}}$ on an unlabeled target dataset $\mathcal{D}$, keeping $\theta_{\mathrm{frozen}}$ fixed: 
\begin{equation}
     \min_{\theta_{\mathrm{adapt}}} \; \mathcal{L}_{\mathrm{SSL}}(\theta_{\mathrm{frozen}},\, \theta_{\mathrm{adapt}};\, \mathcal{D})
\end{equation}
This strategy retains the generic signal representations learned during pretraining in lower layers while enabling domain-specific calibration in the final layer.

\textbf{Downstream Evaluation via Linear Probing:} To evaluate representation quality, we employ linear probing~\cite{al2024pretrainingtorr_ssl,chen2020simple,radford2021clip} on a labeled downstream dataset $\mathcal{D}_{\mathrm{task}}$ containing EEG segment-class pairs $(\mathbf{x},\mathbf{y})$. For each EEG window $\mathbf{x}$, we extract representations from our pretrained encoder $f_{\theta^{*}}$ and train only a linear classifier $g_{\phi}$ to obtain $g_{\phi}(f_{\theta^{*}}(\mathbf{x}))$. Overall, the linear head is adapted by optimizing only for $\phi$, keeping $\theta^{*}$ frozen as defined below:
\begin{equation}
 \phi^{*} 
=
\arg\min_{\phi}\;
\mathbb{E}_{(\mathbf{x},\mathbf{y})\sim \mathcal{D}_{\mathrm{task}}}
\Big[
\mathcal{L}_{\mathrm{task}}\big(g_{\phi}(f_{\theta^{*}}(\mathbf{x})),\mathbf{y}\big)
\Big],
\label{eq:linear_probe}
\end{equation}
where $\mathcal{L}_{\mathrm{task}}$ is the task-specific cross entropy loss.

\subsection{Normalized Evaluation}
To correctly assess the impact of data diversity, we must isolate its effects from variations in computational cost and hyperparameter optimization. Following established protocols for fair comparison under varying data regimes~\cite{al2024pretrainingtorr_ssl}, we use: (1) a fixed computational budget to normalize computation across experiments, and (2) data diversity decomposed with respect to the number of unique EEG windows and (3) unique patients to characterize what the model observes or learns from, during the SSL adaptation.

\textbf{Computational Budget and Data Diversity:} We standardize adaptation compute using the number of optimizer update steps. For an unlabeled training set of data with size $N_{\mathrm{total}}$, batch size $B$, and a chosen reference epoch-equivalent $\mathcal{E}$ for SSL, we define the steps per epoch as:
\begin{equation}
S_{\mathrm{epoch}}=\left\lceil \frac{N_{\mathrm{total}}}{B}\right\rceil,\qquad
S = \mathcal{E}\cdot S_{\mathrm{epoch}},\qquad
C = S\cdot B,
\end{equation}
where $S$ is the total number of optimizer updates and $C$ is the computational budget, representing the total number of EEG windows ``seen'' during training. Equivalently, $C = N \times \mathcal{E}$, where $N$ is the number of unique EEG windows in the adaptation set $\mathcal{D}$. The \textbf{repetition factor $r$}, defined as $C/N$ or $\mathcal{E}$ quantifies how many times, on average, each unique window is processed during training. A model adapted with large $r$ (small $N$, many epochs) sees limited diversity with high repetition, and vice versa.

For each dataset, we vary sample sizes $N \le N_{\mathrm{total}}$ (selecting roughly log-spaced fractions of the total data, rounded to multiples of batch size for stability), while adjusting $\mathcal{E}$ to maintain a fixed $C$. This protocol enables us to observe \emph{how much unique data is needed to achieve peak performance under resource constraints}, allowing us to disentangle whether performance gains arise from increased repetition $r$ (more epochs on limited data) or increased diversity (exposure to more unique samples).

\textbf{Patient vs. Window Diversity.} EEG datasets exhibit hierarchical structure: each patient contributes multiple temporal windows. To determine whether adaptation benefits more from increasing the number of patients or collecting longer recordings, we fix both total windows $N$ and compute budget $C$ while varying the number of patients $P$. For each value of $P$, we randomly sample $P$ patients and extract windows from their available recordings such that the total number of windows $\sum_{i} W_{i} = N$ (where $ W_{i}$ is the number of windows from patient $i$, constrained by that patient's available data). Since $N$ is fixed, smaller $P$ implies more windows per patient on average (longer recordings from fewer patients), while larger $P$ implies fewer windows per patient (brief recordings from many patients). We repeat each configuration with 3 data sampling seeds.

\begin{table}[t]
\centering
\caption{Performance comparison between linear probing (LP) versus our parameter-efficient SSL adaptation (SA). Results shown as mean $\pm$ SEM across 5 seeds. Blue subscripts indicate absolute improvement; best results in \textbf{bold}. Metrics: Balanced Accuracy / Cohen's Kappa / Weighted F1 (TUEV); Balanced Accuracy / AUCPR / AUCROC (TUAB, CHB-MIT). $\theta_{\mathrm{adapt}}$: adapted parameters (of total $\theta$ in EEG-FM). }
\label{tab:optimized_results}
\setlength{\tabcolsep}{2.5pt}
\scalebox{0.8}{
\begin{tabular}{@{}llccc@{}}
\toprule
\textbf{Dataset} & \textbf{Method} & \textbf{Bal.\ ACC} & \textbf{AUCPR / Kappa} & \textbf{AUCROC / F1} \\
\midrule
\multicolumn{5}{@{}l}{\textbf{BIOT} \textcolor{gray}{\small ($|\theta|{=}3.19\text{M}$, $|\theta_{\mathrm{adapt}}|{=}0.92\text{M}$)}} \\[2pt]
TUEV & LP & 30.52 $\pm$ 1.21 & 33.86 $\pm$ 2.84 & 65.87 $\pm$ 1.64\\
     & SA  & \textbf{ 31.78 $\pm$ 0.46} $_{\color{blue}+1.26}$ & \textbf{40.45 $\pm$ 1.41} $_{\color{blue}+6.59}$ & \textbf{69.43 $\pm$ 0.63} $_{\color{blue}+3.56}$ \\[2pt]
TUAB & LP & 63.93 $\pm$ 0.15 & 74.97 $\pm$ 0.13 & 75.08 $\pm$ 0.11 \\
     & SA &\textbf{70.86 $\pm$ 0.23} $_{\color{blue}+6.93}$ & \textbf{77.39 $\pm$ 0.24} $_{\color{blue}+2.42}$ & \textbf{78.34 $\pm$ 0.26} $_{\color{blue}+3.26}$ \\[2pt]
CHB-MIT & LP & 50.74 $\pm$ 0.21 & 1.53 $\pm$ 0.03 & 46.93 $\pm$ 0.30 \\
       & SA & \textbf{52.71 $\pm$ 0.72} $_{\color{blue}+1.97}$ &\textbf{ 31.52 $\pm$ 1.21} $_{\color{blue}+30.0}$ & \textbf{83.08 $\pm$ 1.89} $_{\color{blue}+36.2}$ \\
\midrule
\multicolumn{5}{@{}l}{\textbf{CBraMod} \textcolor{gray}{\small ($|\theta|{=}4.92\text{M}$, $|\theta_{\mathrm{adapt}}|{=}0.44\text{M}$)}} \\[2pt]
TUEV & LP & 28.97 $\pm$ 0.30 & 35.82 $\pm$ 0.94 & 67.27 $\pm$ 0.42 \\
     & SA &  \textbf{44.36 $\pm$ 1.28} $_{\color{blue}+15.4}$ & \textbf{40.91 $\pm$ 1.02} $_{\color{blue}+5.09}$ & \textbf{69.06} $\pm$\textbf{ 0.62} $_{\color{blue}+1.79}$ \\[2pt]
TUAB & LP & 59.36 $\pm$ 0.04 & 56.44 $\pm$ 0.05 & 62.72 $\pm$ 0.07 \\
     & SA & \textbf{66.37 $\pm$ 1.70} $_{\color{blue}+7.01}$ & \textbf{77.16 $\pm$ 0.40} $_{\color{blue}+20.7}$ & \textbf{77.43 $\pm$ 0.52} $_{\color{blue}+14.7}$ \\[2pt]
CHB-MIT & LP & 57.87 $\pm$ 0.20 & 17.35 $\pm$ 0.24 & 54.14 $\pm$ 1.39 \\
       & SA & \textbf{61.37 $\pm$ 3.50} $_{\color{blue}+3.50}$ & \textbf{24.62 $\pm$ 2.31} $_{\color{blue}+7.27}$ & \textbf{83.67 $\pm$ 1.36} $_{\color{blue}+29.5}$  \\
\bottomrule
\vspace{-10mm}
\end{tabular}}
\end{table}

\section{Experiments and Results}
\subsection{Dataset and Setup}
\emph{Data and Preprocessing:} We evaluate on three clinical EEG tasks: (i)~abnormality detection using TUH Abnormal EEG Corpus (TUAB)~\cite{lopez2015tuab}, (ii)~event type classification using TUH EEG Events (TUEV)~\cite{harati2015tuev}, and (iii)~seizure detection in pediatric epilepsy patients using CHB-MIT~\cite{shoeb2009chbmit}. TUAB and TUEV originate from Temple University Hospital and share recording infrastructure with CBraMod's pretraining corpus (TUEG), making them \emph{in-distribution} (ID) for CBraMod and \emph{out-of-distribution} (OOD) for BIOT. CHB-MIT is a pediatric dataset recorded at a different institution with different hardware, making it OOD for both models. For TUAB and TUEV we use the official train/test splits; for CHB-MIT (24 patients) we use patients 1--20 for training, 21--22 for validation, and 23--24 for testing.

We use the common 16 bipolar montage channels in the international 10-20 system following BIOT~\cite{yang2023biot} to obtain clean and uniformly formatted data. Signals are band-pass filtered (0.3--75\,Hz), notch-filtered (60\,Hz), and resampled to 200\,Hz. For SSL adaptation we segment the EEG signal into 30-second non-overlapping windows for all three datasets used in this study; while for downstream evaluation we use 10-second windows (TUAB, CHB-MIT) and 5-second windows (TUEV), following CBraMod~\cite{wang2025cbramod}. 

\emph{Foundation Models and SSL Adaptation:}
We evaluate two EEG foundation models with distinct pretraining objectives: BIOT~\cite{yang2023biot}, a contrastive learning based model pretrained on a sleep EEG corpus (SHHS~\cite{quan1997shhs}) and a proprietary resting EEG dataset (PREST), and CBraMod~\cite{wang2025cbramod}, a masked reconstruction model pretrained on Temple University EEG data (TUEG)~\cite{obeid2016tueg}. We establish baselines through linear probing (LP), where all encoder layers remain frozen and only a linear classification head is trained on labeled downstream data, evaluating off-the-shelf representation quality~\cite{radford2021clip}. 


\emph{Training Details:} For SSL Adaptation, we use only the \emph{training set} of each target dataset using AdamW optimizer~\cite{loshchilov2019decoupled} with learning rate $3 \times 10^{-5}$, weight decay $5 \times 10^{-2}$, and $(\beta_1, \beta_2) = (0.9, 0.999)$, with a cosine annealing schedule (minimum learning rate $10^{-6}$). For linear probe evaluation, we train the classifier for 10 epochs, with learning rate $4 \times 10^{-4}$. We use a batch size of 128 for TUAB and CHB-MIT, and 256 for TUEV. 

\emph{Evaluation Metrics:} For binary tasks (TUAB, CHB-MIT) we report Balanced Accuracy, AUCPR, and AUCROC, with AUCROC as the monitor score. For multi-class classification (TUEV, 6 classes) we report Balanced Accuracy, Cohen's Kappa, and Weighted F1, with Kappa as the monitor score. All results are reported as mean $\pm$ standard error of the mean (SEM) over 5 independent random seeds, on the test set, if not mentioned otherwise.

\begin{figure}[!t]
\centering
\includegraphics[width=1.0\textwidth]{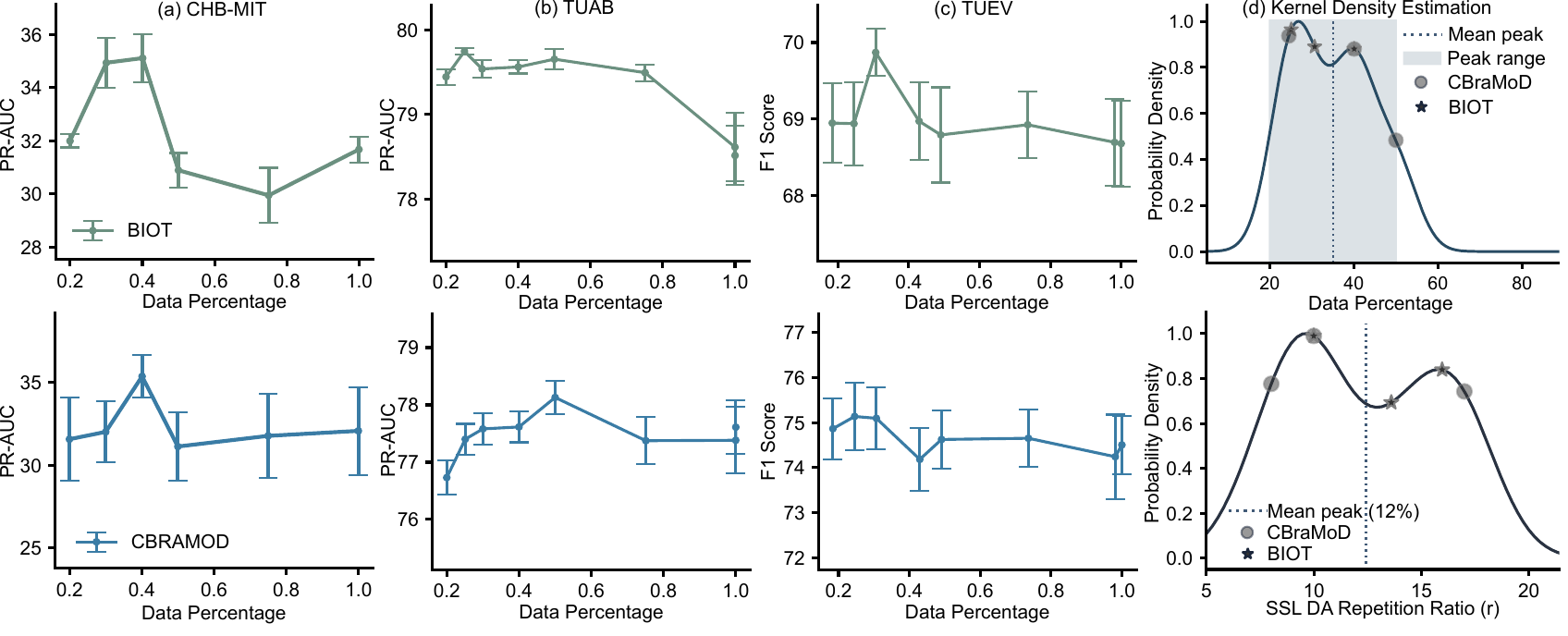}
\caption{SSL adaptation exhibits data efficiency under normalized compute $C{=}N{\times}\mathcal{E}$: performance peaks at $20-50$\% of data then plateaus. (a) CHB-MIT seizure detection (AUCPR), $N_{\text{total}}{=}4{,}174$; (b) TUAB abnormality detection (AUCPR), $N_{\text{total}}{=}57{,}614$; (c) TUEV event classification (F1), $N_{\text{total}}{=}86{,}587$. (d) Kernel density estimation confirms saturation at $N/N_{\text{total}}{=}0.2$--$0.5$ (repetition factor $r{=}\mathcal{E}{=}12$). Green: BIOT; Blue: CBraMod. Error bars denote SEM (3 data shuffle ${\times}$ 3 model seeds).
} \label{fig:divN}
\end{figure}

\begin{figure}[!ht]
\centering
\includegraphics[width=0.8\textwidth]{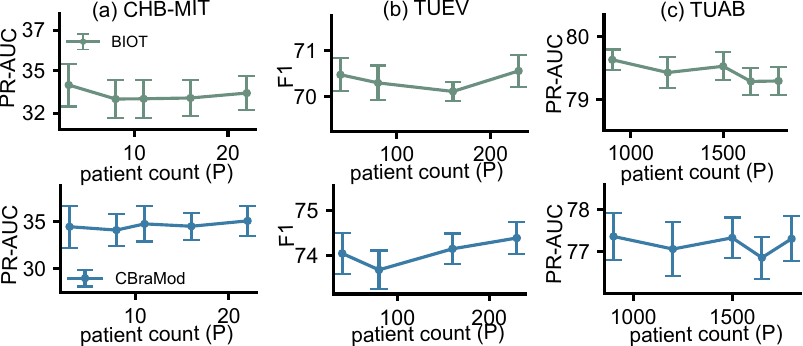}
\caption{SSL adaptation performance is patient-agnostic under fixed computational budget $C$. At constant $N{=}P{\times}W$, varying patient count $P$ yields flat performance curve: (a) CHB-MIT seizure detection (AUCPR), $P{\in}[3,8,11,16,22]$; (b) TUEV event classification (F1), $P{\in}[40,80,160,230]$; (c) TUAB abnormality detection (AUCPR), $P{\in}[900,1200,1500,1650,1800]$. Green: BIOT; Blue: CBraMod. Error bars denote SEM over 3 data shuffle ${\times}$ 3 model initialization seeds.} \label{fig:main_divP}
\end{figure}

\subsection{Parameter-Efficient SSL Adaptation on Clinical Tasks}
Table~\ref{tab:optimized_results} summarizes the comparison between EEG-FM linear probing (LP) and our SSL adaptation. Across all three clinical datasets, SSL adaptation (SA) improves performance ranging from $+1.79$ weighted F1 score (CBraMod on TUEV) to $+36.2$  AUCROC (BIOT on CHB-MIT). Importantly, SA updates only $0.92$M of $3.19$M parameters for BIOT ($28.8\%$) and $0.44$M of $4.92$M for CBraMod ($9.0\%$). Thus, we obtain better aligned feature representations using parameter efficient adaptation regardless of pretraining objective. For BIOT, which is OOD on all three datasets, adaptation yields $+6.59$ Kappa on TUEV, $+3.26$ AUCROC on TUAB ($63.93\to70.86\%$ Balanced Accuracy), and $+36.2$ AUCROC on CHB-MIT seizure detection ($46.93\to83.08\%$ AUCROC). For CBraMod, gains are observed on both ID tasks (TUAB: $+14.7$ AUCROC; TUEV: $+5.09$ Kappa) and the OOD task (CHB-MIT: $+29.5$ AUCROC: ($54.14\to83.67\%$)). These improvements indicate that pretrained representations are not sufficiently aligned to target-domain signal characteristics for direct deployment. The gains are particularly pronounced on OOD tasks, similar to findings in medical imaging where SSL-based adaptation of pretrained models yields the largest improvements under distribution shifts~\cite{azizi2023robust}.

CHB-MIT seizure detection task, a classical rare event with prevalence $1.4\%$, exposes a critical limitation of frozen foundation model features: EEG-FM linear probing achieves near-random AUCPR performance (BIOT: 1.53\%, CBraMod: 17.35\%), barely exceeding random classifier baseline precision (1.48\%). We verified this result across all 5 seeds, confirming that frozen features fail to capture discriminative patterns for rare clinical events. SSL adaptation recovers clinically meaningful performance (BIOT: 31.52\%, CBraMod: 24.62\% AUCPR), demonstrating that even modest unlabeled target-domain training can align representations when frozen features provide no discriminative signal.

To investigate the trade-off between adaptation capacity and parameter efficiency, we selectively update different encoder layer subsets during SSL adaptation (final layer only, final two layers, or all layers). As shown for the CHB-MIT dataset in Table~\ref{tab:layer_ablation}, we observe only marginal gains (+1.17\% AUCPR) for BIOT and even slightly degraded performance for CBraMod (31.25\% AUCPR $\to$ 29.41\%), at the cost of significantly more parameters. Final-layer adaptation, thus, provides an optimal cost-performance balance. This is consistent with transfer learning principles 
that selective updates preserve pretrained knowledge while enabling domain-specific calibration~\cite{yosinski2014transferable}.

\textbf{Takeaway:} \emph{Across all datasets, models and pretraining objectives evaluated, last-layer SSL adaptation yields consistent gains (up to $20\times$ AUCPR), making it a highly efficient and lightweight adaptation method especially in cases where distribution shifts impact performance.}

\subsection{Data Efficiency under Normalized Compute Budget:} We systematically vary the amount of unique unlabeled data ($N$) while holding total compute ($C{=}N{\times}\mathcal{E}$) constant across all adaptation runs. Figure~\ref{fig:divN} shows downstream performance as a function of data percentage for both models across all three datasets. Interestingly, both BIOT and CBraMod reach peak performance using only $20-50$\% of the available unlabeled data across all tasks. 
On CHB-MIT, both models achieve ${\approx}$35\% AUCPR at 40\% of the data ($N{=}86$K, repetition factor $r{=}12$). On TUEV, both peak at ${\approx}$30\% with $r{=}14$, then plateau. The kernel density plot (Fig.~\ref{fig:divN}d), aggregated over all models, datasets, and seeds, confirms that the distribution of peak performance concentrates at $N/N_{\text{total}}{=}0.2$--$0.5$. Beyond this point, additional unique data provides no measurable benefit under fixed compute. The model sees each window fewer times without gaining sufficient new information to offset the reduced repetition. This is consistent with recent findings on pretraining data diversity for SSL in vision, where beyond a saturation threshold, increasing unique samples under fixed compute yields diminishing returns as models require sufficient repetition to consolidate learned features~\cite{al2024pretrainingtorr_ssl}. We note that this result characterizes the optimal diversity–repetition balance at a given compute budget. 

\textbf{Takeaway:} \emph{Under fixed compute, peak SSL adaptation requires only $20-50$\% of available unlabeled data with sufficient repetition of approx. $12$, enabling rapid deployment with minimal data collection.}
 
\begin{table}[t]
\centering
\scriptsize
\caption{Layer unfreezing ablation on CHBMIT dataset with one seed. Unfreezing only the final layer ($L$) provides optimal parameter efficiency.}
\label{tab:layer_ablation}
\small
\scalebox{0.8}
{
\begin{tabular}{@{}llcccc@{}}
\toprule
Model & Layers& Param. & AUCROC & AUCPR & Bal. Acc \\
\midrule
\multirow{3}{*}{BIOT}    & $L$ & 0.92M &88.67 & 35.62 & 60.62 \\
                         & $L,L-1$& 1.71M & 88.44 & 35.05 & 60.82 \\
                         & All  & 3.19M  & 88.60 & 36.79 & \textbf{62.18} \\
\midrule
\multirow{3}{*}{CBRAMOD} & $L$ & 0.44M & 87.16 & 31.25 & \textbf{67.60} \\
                         & $L,L-1$ & 0.84M & 86.62 & 21.90 & 58.69 \\
                         & All  & 4.4M  & 88.20 & 29.41 & 50.98 \\
\bottomrule
\end{tabular}}
\end{table}
\subsection{Total EEG Window Count Dominates Over unique patient ID:}
To disentangle whether adaptation benefits more from patient diversity
or temporal coverage, we fix total windows $N{=}P{\times}W$ and compute
$C$, then systematically vary the patient-window composition.
Figure~\ref{fig:main_divP} shows downstream performance as a function
of patient count $P$ across all three datasets. Across the tested ranges
(CHB-MIT: $P{\in}\{3,8,11,16,22\}$;
TUEV: $P{\in}\{40,80,$ $160,230\}$;
TUAB: $P{\in}\{900,$ $1200, 1500,$
$1650, 1800\}$),
increasing number of unique patients, produces no significant change in downstream performance, suggesting that SSL objectives capture transferable temporal structure rather than patient-specific features.

\textbf{Takeaway:} \emph{Under FM-initialized SSL adaptation with fixed samples and compute, total windows seen, not patient count, affects the performance, suggesting clinical sites can adapt effectively using longer recordings from smaller cohorts.}

\section{Conclusion}
We present a parameter-efficient SSL adaptation strategy for EEG foundation models that updates 9\% of parameters, yet achieves substantial performance gains. Across three clinical tasks and two models with distinct architectures and pretraining objectives, 
adaptation consistently outperforms linear probing (AUCROC gains up to +36.2 points). On rare-event seizure detection (CHB-MIT), adaptation recovers clinically meaningful performance (31.52\% AUCPR) from near-random baselines (1.53\%), demonstrating that domain adaptation is essential before clinical deployment. We systematically characterize data requirements under fixed compute, finding 
that peak performance requires only $20-50$\% of available data. Additionally, with fixed compute budget and total samples while varying patient-window composition ($N{=}P{\times}W$), no significant performance difference is observed, suggesting that for FM-initialized SSL adaptation, temporal window coverage may matter more than patient diversity. We note that for one of the datasets (CHB-MIT), this observation is over limited patient range (3--22). Future work should compare alternative parameter-efficient pretraining methods,  vary the compute budget $C$ within our fixed-compute design to test
the robustness of findings in this work across compute regimes, and test whether these data-efficiency patterns hold for general-purpose time-series foundation models and other medical time-series data.

\begin{credits}
\subsubsection{\ackname} This work was supported by the Else Kröner Fresenius Foundation, German Research Foundation (DFG): 493665037, SPP 2241 - PN 520287829, the Machine Learning Cluster of Excellence EXC number 2064/1 PN 390727645, Tübingen AI Center and BMFTR (01GQ2502). M.D. is a member of the International Max Planck Research School for Intelligent Systems Tübingen (IMPRS-IS). The authors thank C3N lab members, Prof. Dr. Jakob H. Macke and Julius Vetter for discussion and feedback. 

\subsubsection{\discintname}
The authors declare no competing interests.
\end{credits}
%
%
%
\bibliographystyle{splncs04}
\bibliography{egbib}

\end{document}